%% file: iclr2026-conference-shorter.tex
\documentclass{article} 
\usepackage{iclr2026_conference,times}

\input{math_commands.tex}

\usepackage{hyperref}
\usepackage{url}

\usepackage{amsmath}
\usepackage{multirow}
\usepackage{multicol}
\usepackage{graphicx}
\usepackage{wrapfig}
\usepackage{caption}

\usepackage{algorithm}
\usepackage{algorithmic}

\title{Labeled TrustSet Guided: Batch Active Learning with Reinforcement Learning}

\author{Guofeng Cui, Yang Liu, Pichao Wang, Hankai Hsu, Xiaohang Sun, Xiang Hao \& Zhu Liu \thanks{Guofeng Cui and Pichao Wang are now with Nvidia. This work was done while they were at Amazon} \\
Prime Video, Amazon \\
Seattle, USA \\
\texttt{\{gfcui404, pichaowang\}@gmail.com} \\
\texttt{\{yangnliu, hankhsu, sunking, xianghao, zhuzliu\}@amazon.com} \\
}

\iclrfinalcopy 
\begin{document}

\maketitle

\begin{abstract}
Batch active learning (BAL) is a crucial technique for reducing labeling costs and improving data efficiency in training large-scale deep learning models. Traditional BAL methods often rely on metrics like Mahalanobis Distance to balance uncertainty and diversity when selecting data for annotation. However, these methods predominantly focus on the distribution of unlabeled data and fail to leverage feedback from labeled data or the model’s performance. To address these limitations, we introduce TrustSet, a novel approach that selects the most informative data from the labeled dataset, ensuring a balanced class distribution to mitigate the long-tail problem. Unlike CoreSet, which focuses on maintaining the overall data distribution, TrustSet optimizes the model’s performance by pruning redundant data and using label information to refine the selection process. To extend the benefits of TrustSet to the unlabeled pool, we propose a reinforcement learning (RL)-based sampling policy that approximates the selection of high-quality TrustSet candidates from the unlabeled data. Combining TrustSet and RL, we introduce the \textbf{B}atch \textbf{R}einforcement \textbf{A}ctive \textbf{L}earning with \textbf{T}rustSet (\textbf{BRAL-T}) framework. BRAL-T achieves state-of-the-art results across 10 image classification benchmarks and 2 active fine-tuning tasks, demonstrating its effectiveness and efficiency in various domains.
\end{abstract}

\section{Introduction}
In the era of deep learning, large-scale labeled datasets are indispensable for training models on complex tasks. Active learning (AL) provides an efficient approach to reduce the labeling costs by intelligently selecting critical subsets from unlabeled data for annotation~\citep{zhan2022comparative,yang2024plug,nemeth2024compute,safaei2024entropic}. Batch active learning (BAL)~\citep{citovsky2021batch}, a variant of AL, further improves this process by selecting data points in groups (batches), thereby reducing the overhead associated with model retraining and oracle interactions.

In most modern BAL methods, the selection strategy is typically based on two factors: uncertainty and diversity. Uncertainty-based methods focus on choosing the most ambiguous or difficult data, which is likely to improve the model, but this often results in selecting redundant data that doesn't sufficiently cover the data distribution~\citep{shen2017deep}. On the other hand, diversity-based methods aim to ensure a representative subset by covering as many different types of data as possible, but they may neglect critical uncertain samples near the decision boundaries. For instance, CoreSet~\citep{phillips2017coresets} selects subsets that reflect the overall data distribution, ensuring diversity by minimizing the distance between the selected subset and the full dataset. While methods like Cluster-Margin~\citep{citovsky2021batch} combine diversity and uncertainty to improve data selection, they still have limitations, such as overlooking feedback from the labeled dataset, ignoring class distribution, and potentially inheriting the long-tail distribution problem.

To address these challenges, we propose TrustSet, a novel data selection approach that distinguishes itself from CoreSet by emphasizing the utilization of label information. TrustSet focuses not only on ensuring diversity but also on selecting data that is most beneficial for improving the model's performance and releasing class imbalance problem. TrustSet differs from CoreSet in two ways:

\textbf{Objective}: TrustSet is designed to optimize the model's performance, with an explicit focus on improving accuracy and tackling the long-tail distribution problem by selecting crucial data that has a high potential to be forgotten by the model~\citep{toneva2018empirical}. In contrast, CoreSet focuses on representing the full data distribution without directly considering the impact on the model’s learning process, which can lead to inheriting undesirable distributional imbalances.

\textbf{Data Source}: TrustSet leverages labeled data, utilizing ground truth labels to prune redundant and noisy data and ensure that the selected subset is balanced across classes. This approach contrasts with CoreSet, which selects data purely from unlabeled pool, without considering feedback from trained model. As a result, CoreSet-based methods can miss the opportunity to incorporate critical information about model’s current performance, potentially leading to suboptimal data selections.

TrustSet's balanced class distribution ensures better handling of the long-tail distribution problem, where underrepresented classes are more likely to be included in the training process. To construct TrustSet, we use the GradNd method~\citep{paul2021deep}, which ranks data based on the gradient norms of model updates, prioritizing data points that contribute most to model learning. Furthermore, to improve the data quality, we incorporate SuperLoss~\citep{castells2020superloss}, which follows a curriculum learning strategy to assign higher importance to easier data in early training stages, while still considering difficult samples later.

\begin{figure}
  \begin{center}
    \includegraphics[width=.7\textwidth]{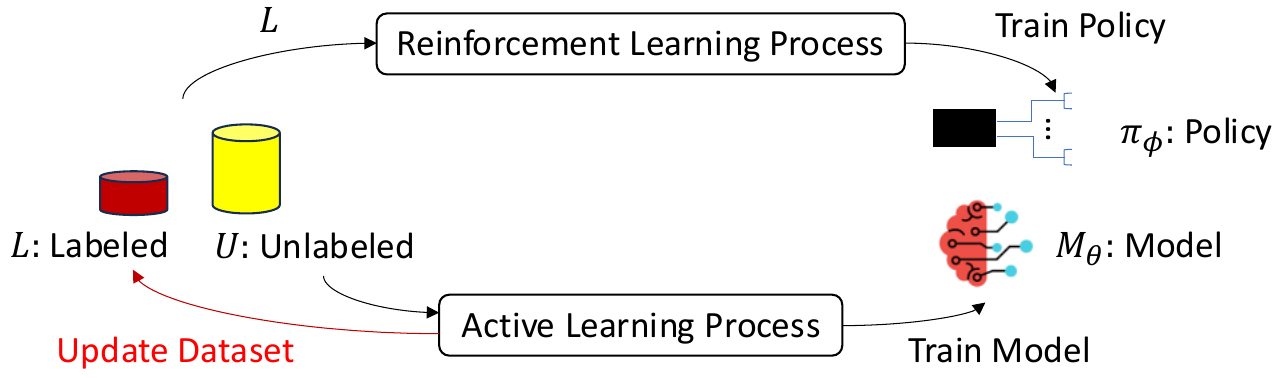}
  \end{center}
  \caption{An overview of the BRAL-T framework, which consists of two main processes: (1)\textbf{Active Learning} for data selection. (2)\textbf{Reinforcement Learning} to approximate TrustSet selection.}
  \label{fig:general}
\end{figure}


However, extending TrustSet to the unlabeled data pool presents a challenge, as the selection process requires label information. To overcome this, we introduce an RL-based policy for approximating the selection of high-potential TrustSet candidates from the unlabeled data pool. Unlike previous RL-based active learning approaches, which often require frequent retraining of the model and rely heavily on complex reward structures~\citep{fang2017learning,zhang2023algorithm}, our method minimizes retraining costs by leveraging TrustSet to guide the reinforcement learning process.

To this end, we propose a novel batch active learning framework called BRAL-T (\textbf{B}atch \textbf{R}einforcement \textbf{A}ctive \textbf{L}earning with \textbf{T}rustSet extraction), which integrates TrustSet and RL-based policies for efficient data selection. The framework consists of two primary components: (1) TrustSet extraction, which ensures that the labeled dataset contributes optimally to model performance and maintains a balanced class distribution, and (2) RL-based subset selection, where a learned policy selects from the unlabeled data pool to approximate TrustSet. This significantly reduces the need for repeated oracle queries and model retraining. As shown in Figure~\ref{fig:general}, BRAL-T is implemented with two processes: reinforcement learning (RL) for policy training and active learning (AL) for model training.

Our contributions are summarized as following: (1) We introduce TrustSet, a novel method for data selection that leverages label information to balance uncertainty, diversity, and class distribution, thus releasing class imbalanced issue and improving performance of AL. (2) We develop an RL-based data selection policy that bridges the gap between TrustSet’s label dependency and the unlabeled setting of active learning, allowing for more efficient and targeted data selection. (3) We propose BRAL-T, a new batch active learning framework that integrates TrustSet and RL to reduce the computational burden of active learning while improving model performance. We demonstrate that BRAL-T achieves state-of-the-art performance across multiple image classification and active fine-tuning tasks.


\section{Related Work}

\textbf{Active Learning:} Active learning aims to reduce labeling costs by selecting informative data based on uncertainty and diversity. \citet{shen2017deep} explored uncertainty-based methods for Named Entity Recognition, later integrating diversity. Galaxy~\citep{zhang2022galaxy} leveraged a model confidence graph to estimate uncertainty, while \citet{yuan2020cold} applied self-supervised models for diversity-based selection. Recent works~\citep{liu2019deep,ash2019deep,sinha2019variational,margatina2021active,ash2021gone,citovsky2021batch,kim2021task,gentile2022achieving} highlight the uncertainty-diversity trade-off, where uncertainty-focused methods risk redundancy, and diversity-based methods may overlook critical uncertain samples. However, the batch sampling methods in these works tend to ignore the distribution of selected data, leading to further redundancy

\textbf{Batch Active Learning:} Active learning methods, aiming to minimize oracle queries, prefer batch data processing over individual sample handling. Batch active learning approaches~\citep{zhang2023algorithm,ash2021gone,kirsch2019batchbald,citovsky2021batch,sener2017active} concentrate on batch sampling to reduce costs and preserve subset distribution.  BatchBald~\citep{kirsch2019batchbald} addressed the lack of joint informativeness in batch sampling with an entropy-based selection. Cluster-Margin~\citep{citovsky2021batch} used Hierarchical Agglomerative Clustering to identify highly uncertain samples at scale. Despite their success in computer vision tasks, existing methods overlook feedback from selected samples, such as accuracy changes, which could be crucial for refining data sampling strategies.

\textbf{Active Learning with RL:} RL has been explored to learn data selection policies~\citep{zhang2023algorithm,fang2017learning,liu2019deep,gong2022meta,smit2021medselect,casanova2020reinforced}. Some approaches\citep{fang2017learning,gong2022meta,smit2021medselect,casanova2020reinforced} defined rewards based on target model metrics (e.g., accuracy, AUROC), requiring frequent retraining and suffering from credit assignment issues. Instead, \citet{liu2019deep} used the Mahalanobis distance for reward definition, while TAILOR~\citep{zhang2023algorithm} formulated class balance as a reward signal. However, the former is task-specific (Re-ID), and the latter does not directly align with model accuracy. To address these limitations, we propose an RL-based active learning approach that avoids retraining while maintaining high correlation with task performance.

\section{Problem Definition}
\label{sec: definition}

In this section, we formally define the active learning problem in batch setting, following \citet{sener2017active}, and introduce the TrustSet selection problem. We consider an $\mathbb{C}$-class classification task over a compact input space $\mathcal{X}$ and a label space $\mathcal{Y}=\{1,\dots,\mathbb{C}\}$. Our goal is to train a target model $M_\theta$ with parameters $\theta$ to optimize a loss function $\ell(M_\theta(\mathbf{x}),y): \mathcal{X}\times \mathcal{Y}\rightarrow R$, where $\hat{y}=M_\theta(\mathbf{x})$ refers to the predicted category by the target model and cross entropy loss is widely used as $\ell$ in classification task. In practice, we assume a large collection of data points sampled i.i.d. over the space $\mathcal{Z}=\mathcal{X}\times \mathcal{Y}$ as $\{\mathbf{x}_k, y_k\}_{k\in[n]}\sim p_\mathcal{Z}$, where $n$ refers to the total amount of data points and $[n]=\{1,2,\dots,n\}$.

For active learning problem, we further define the labeled dataset with $|L|$ data points as $L=\{\mathbf{x}_k, y_k\}^{|L|}_k$ and the unlabeled data pool with $|U|$ data points as $U=\{\mathbf{x}_k\}^{|U|}_k$. In general, $|L|\ll|U|$ and $L\cap U = \emptyset$. For the $i$-th active learning iteration, the labeled and unlabeled dataset are defined as $L_i$ and $U_i$ respectively. We aim to select a data subset $S_{U_i}\subseteq U_i$ to be labeled by an oracle and added to the current labeled dataset $L_i$ forming an enhanced labeled dataset $L_{i+1}=L_i\cup S_{U_i}$. We abbreviate $S_{U_i}$ as $S_i$ for readability in the rest of the paper. Training on $L_{i+1}$, the model $M_{\theta_{L_{i+1}}}$ with the trained parameter $\theta_{L_{i+1}}$ is expected to achieve the best performance across all possible choices of $S_i$ which can be formulated as the following Eq.~\ref{eq:aim}:
\begin{equation}
\label{eq:aim}
    S^*_i = \argmin_{S_i \subseteq U_i: |S_i|\leq b} E_{\mathbf{x},y\sim p_{\mathcal{Z}}}[\ell(M_{\theta_{L_{i+1}}}(\mathbf{x}), y)]
\end{equation}
where $S^*_i$ refers to the optimal data subset for active learning and $b$ refers to the size of data subset that needs to be selected in each iteration. As a result, Eq.~\ref{eq:aim} indicates that we aim to select a data subset $S_i$ from $U_i$ to enhance $L_i$ such that the trained model $M_{\theta_{L_{i+1}}}$ achieves the minimal loss value.

Directly solving the above optimization problem is challenging due to the large number of possible choice of $S_i$ from $U_i$. To address this, we consider an ideal scenario where label information for $U_i$ is available and analyze the entire dataset $D=L_i\cup U_i$ to identify the most important samples that contribute to model training. Formally, we define the TrustSet $T_D$ to be the important data of $D$ as the following Eq.~\ref{eq:trustset}:
\begin{equation}
\label{eq:trustset}
\begin{split}
    T_{D}= \argmin_{S_D\subseteq D: |S_D|\leq b_T}E_{\mathbf{x},y\sim p_{\mathcal{Z}}}[\ell(M_{\theta_{S_D}}(\mathbf{x}), y)]
    \quad \text{s.t.}\text{  balance}(S_D)
\end{split}
\end{equation}
where $S_D$ refers to a data subset selected from $D$ and we require $S_D$ to be balanced across $\mathbb{C}$ categories for TrustSet selection to alleviate long-tail distribution problem; $M_{\theta_{S_D}}$ refers to the model trained on $S_D$ and $b_T$ refers to the predefined size of TrustSet. Eq.~\ref{eq:trustset} indicates that given the limited size of data subset, $T_{D}$ contains the most useful information from the dataset to train the model. Combining with Eq.~\ref{eq:aim}, we instead select $\Tilde{S_i}$ from the unlabeled data pool $U_i$ for active learning as:
\begin{equation}
\Tilde{S_i} = \argmin\limits_{S_i\subseteq U_i, |S_i|\leq b} d(S_i, T_D\cap U_i)
\end{equation}
where $d(\cdot,\cdot)$ indicates a statistical distance function (e.g. Wasserstein Distance) for two distributions and we aim to select a data subset that has a distribution similar to $T_D\cap U_i$. As $T_D$ contains the most significant data points for model training, enhancing $L_i$ with such data points could also benefit the performance of the trained model. As a result, $\Tilde{S_i}$, which approximates $T_D\cap U_i$, also contributes to a significant improvement of model training. Compared to Eq.~\ref{eq:aim}, TrustSet is more reliable due to the use of annotation labels and the balance requirement releases the class imbalance problem of collected labeled dataset. Additionally, Eq.~\ref{eq:trustset} can be more straightforwardly solved using data pruning methods~\citep{paul2021deep,park2023robust,tan2023data}.

However, label information of $U_i$ is not available under the active learning assumption. To solve this problem, we train a data selection policy $\pi_{\phi_i}$ with RL method on the labeled dataset $L_i$, learning to select $\Tilde{S_i}$ as an approximation of $T_D\cap U_i$. In the $i$-th iteration, we create an environment similar to the active learning setting by randomly sampling two subsets $\mathcal{L}$ and $\mathcal{U}$ from labeled dataset $L_i$, where the labels of $\mathcal{L}$ are retained to simulate labeled dataset while the labels of $\mathcal{U}$ are omitted to simulate unlabeled dataset. The whole dataset in this environment is then defined as $\mathcal{D}=\mathcal{L}\cup \mathcal{U}$, ensuring $\mathcal{L}\cap\mathcal{U}=\emptyset$ and $|\mathcal{L}|\ll|\mathcal{U}|$. Although we omit the labels of $\mathcal{U}$ for active learning purpose, we can still leverage the label to extract $T_\mathcal{D}$. Taking $\mathcal{L}$ and $\mathcal{U}$ as input, $\pi_{\phi_i}$ is trained to select a subset $\mathcal{S}$ from $\mathcal{U}$, with the reward defined as:
\begin{equation}
\label{eq:rl_train}
    R = -d(\mathcal{S}, T_\mathcal{D}\cap \mathcal{U})
\end{equation}
where $\mathcal{S}=\pi_{\phi_i}(\mathcal{L},\mathcal{U})$ and we optimize the parameters $\phi_i$ to minimize the statistical distance between $T_D\cap \mathcal{U}$ and $\mathcal{S}$. In this way, $\pi_{\phi_i}$ learns to select data from the unlabeled dataset that has a high potential to be included in the TrustSet based on the feature space. After training, $\pi_{\phi_i}$ is applied to the real unlabeled data pool $U_{i}$ to select $S_{i}=\pi_{\phi_{i}}(L_{i}, U_{i})$.

\begin{figure}[t]
\centering
\includegraphics[width=.75\columnwidth]{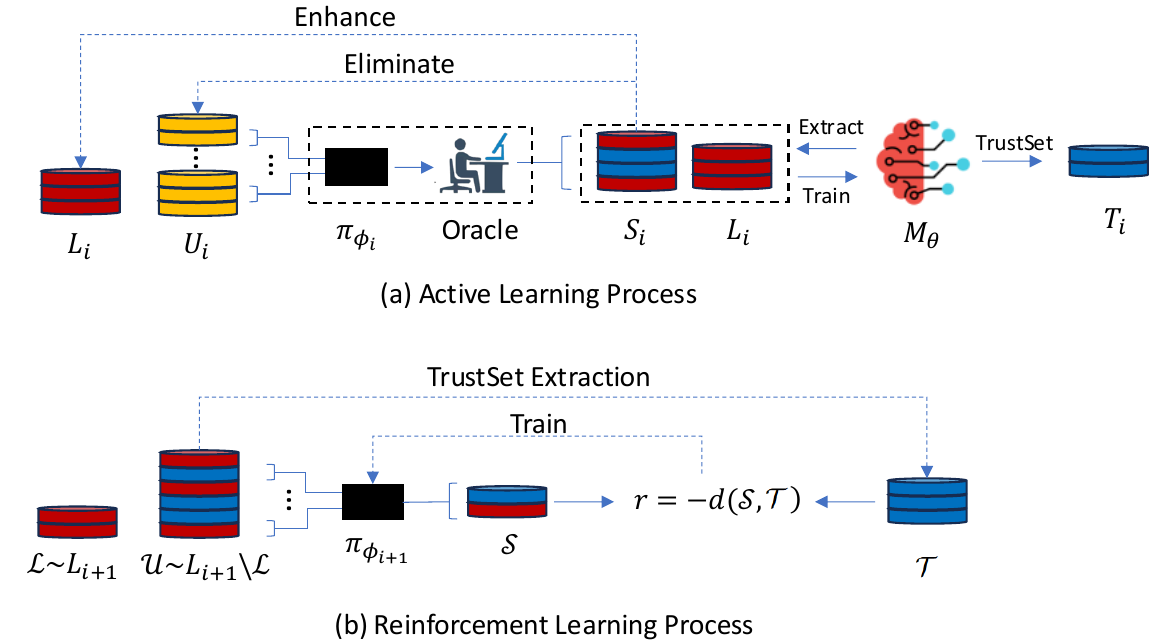}
\caption{Details of BRAL-T. (a) Active Learning Process: $\pi_{\phi_i}$ selects the subset $S_i$ from $U_i$ for the oracle to annotate, and the model $M_{\theta_{L_{i+1}}}$ is trained on $S_i \cup L_i$. (b) RL process: we sample $\mathcal{L}$ and $\mathcal{U}$ from $L_{i+1}$ to train policy $\pi_{\phi_{i+1}}$.}
\label{fig:overview}
\end{figure}

In general, for each active learning iteration, we solve Eq.~\ref{eq:aim} in two processes as shown in Figure~\ref{fig:overview}. \textbf{In the active learning process}, data subset $S_i$ is selected by $\pi_{\phi_i}(L_i, U_i)$ and passed to the oracle for annotation. The model $M_\theta$ is then trained on the enhanced labeled dataset $L_{i+1}$. Moreover, the new unlabeled data pool $U_{i+1}$ is achieved by eliminating $S_i$ from $U_i$ and the TrustSet $T_i$ is extracted from $L_{i+1}$ by solving Eq.~\ref{eq:trustset} to facilitate the next RL process. \textbf{In the RL process}, we create the environment based on $L_{i}$ and train $\pi_{\phi_{i}}$ using the reward function defined in Eq.~\ref{eq:rl_train}.

\section{Method}
\label{sec:method}
In this section, we introduce BRAL-T framework in detail. In Section~\ref{sec:method_trustset}, we introduce a TrustSet construction method based on GradNd score~\citep{paul2021deep}. In Section~\ref{sec:method_rl}, we illustrate the details of the RL module and describe how we use the learned policy to select subsets from the unlabeled pool.

\subsection{TrustSet}
\label{sec:method_trustset}

In general, the TrustSet should retain important data and tend to be class-balanced. However, it is almost impossible to directly solve Eq.~\ref{eq:trustset} due to the large size of labeled dataset and time-consuming of $M_{\theta_{S_D}}$ training. As a result, we introduce a TrustSet extraction method based on the GradNd score~\citep{paul2021deep} by analyzing the performance of model trained on entire trainset $L$ rather than selected subset $S$. This score is defined as the expected value of the gradient norm term with respect to a differentiable model and a data sample $x$:
\begin{equation}
GradNd = E\| \sum_{k=1}^K \nabla_{M^{(k)}_\theta}\ell(M_\theta(x), y)^T \nabla_\theta M^{(k)}_\theta(x) \|
\label{eq: gradnd}
\end{equation}
In this equation, $K$ denotes the number of logits, $M_\theta(x)\in\mathbb{R}^K$ refers to the output of model and $M^{(k)}_\theta(x)$ represents the result of the $k$-th logit from the model $M_\theta$. For instance, in an image classification task, $\ell$ represents the cross-entropy loss, $K=\mathbb{C}$ is the number of categories, and $M^{(k)}_\theta(x)$ is the logit output for the $k$-th category. Data samples that result in a large gradient value tend to contain information that the model has not yet learned, as the model would update significantly based on such data. As demonstrated by the experimental analysis from \citet{paul2021deep}, data with a higher GradNd score tend to be forgotten samples for the target model during training and are more important for further training. However, the GradNd score might lead to a class imbalance problem when the data subset primarily contains difficult images for certain categories. To mitigate the long-tail distribution problem, we sort data by class using the GradNd score and select the top-N data for each category. For the image classification task, we follow \citet{paul2021deep} in omitting the term $\nabla\theta M^{(k)}_\theta(x)$ from Eq~\ref{eq: gradnd} and calculate the EL2N score to approximate GradNd.

\textbf{Curriculum Learning:} Data with high GradNd scores tend to be difficult and uncertain samples. As suggested by previous works \citep{ash2021gone,citovsky2021batch,gentile2022achieving}, a training set focusing on uncertainty could result in high redundancy and fail to train a model that captures general features. We reconsider this issue from another important perspective. Difficult samples contain noise that can interfere with model predictions and increase the difficulty for the model to learn the boundaries between categories. With a limited amount of data, easy examples could help the model capture features and cluster data within the same category. Following the principles of curriculum learning~\citep{tang2019self, castells2020superloss}, we assign larger weights to easier data samples in the early active learning iterations and leverage Super Loss~\citep{castells2020superloss} on top of the task loss $\ell$. For each data sample $(x,y)$, the super loss $\ell_{s}$ is defined as:
\begin{equation}
\ell_{s}(M_\theta(x), y) = (\ell(M_\theta(x), y) - \tau)\sigma + \lambda(\log{\sigma})^2
\end{equation}
where $\tau$ is the threshold for separating easy and hard samples, and $\lambda$ is the weight of the regularization term. Both $\tau$ and $\lambda$ are hyperparameters, while $\sigma$ is learnable and indicates the weight assigned to the task loss. To minimize $\ell_s$, data with task loss $\ell<\tau$ will be assigned a larger weight $\sigma$, and data with $\ell>\tau$ will be assigned a smaller weight $\sigma$. Since model training on uncertain data typically results in larger losses compared to easier data, super loss adaptively adjusts the weight for data samples. Meanwhile, the scale of $\sigma$ is determined by $\lambda$. As $\lambda$ increases, the value of $\sigma$ tends to be 1 and has less effect on the task loss. Specifically, when $\lambda \rightarrow \infty$, $\sigma$ will always be 1 to minimize the regularization term, making $\ell_s$ equal to $\ell - \tau$. As shown in Section~\ref{sec:abl}, with Super Loss, proposed method achieves better performance. In the following sections, $T_D$ refers to the TrustSet with Super Loss unless explicitly stated.

\subsection{Reinforcement Learning}
\label{sec:method_rl}

\begin{figure}[t]
\centering
\includegraphics[width=.8\linewidth]{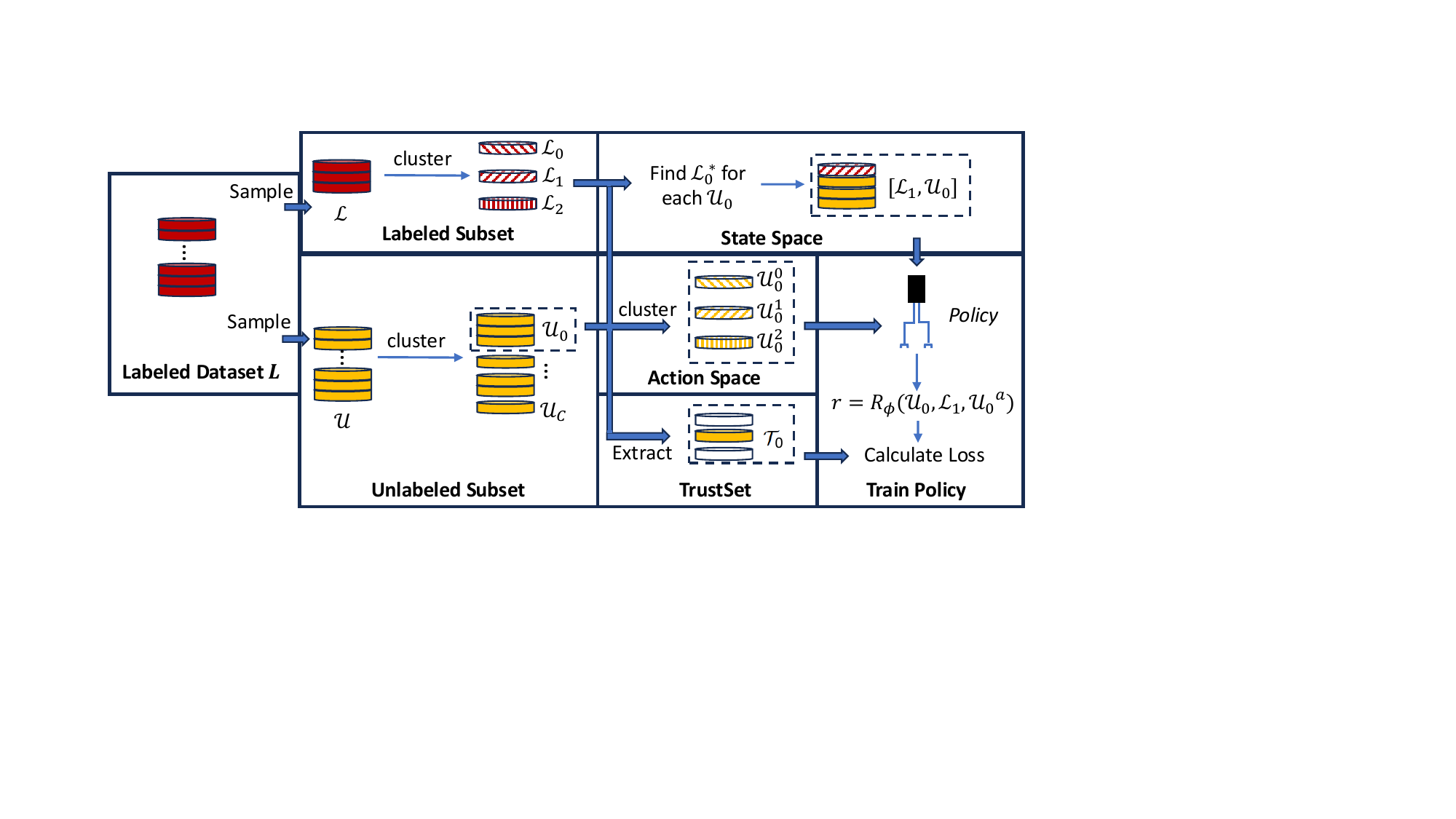}
\caption{Reinforcement learning process. We use $\mathcal{U}_0$ as an example for process in the figure.}
\label{fig:rl}
\end{figure}

The TrustSet is collected with label information to ensure class balance and improved reliability. However, during the active learning process, the policy needs to be applied to select a subset from the unlabeled data pool $U_i$. To address this, we create an environment with similar conditions and apply RL to train a policy for subset selection, where the TrustSet serves as the target subset. In the remainder of this section, we first define the state, action, and reward for the RL task in general, followed by an illustration of the overall process.

\textbf{State $(\mathcal{L}^*_c, \mathcal{U}_c)$.} We randomly sample $\mathcal{L}$ as a labeled dataset and $\mathcal{U}$ as an unlabeled data pool from $L_i$ to train the policy $\pi_{\phi_{i}}$. For each sampled dataset $\mathcal{D}=\mathcal{L}\cup\mathcal{U}$, we extract $T_\mathcal{D}$ and calculate $T_\mathcal{D}\cap \mathcal{U}$ as the target selected subset. For convenience, we define $\mathcal{T}=T_\mathcal{D}\cap \mathcal{U}$. Using all data in $\mathcal{L}$ and $\mathcal{U}$ as input is computationally expensive and challenging for learning an effective policy. It is beneficial to have alternative representations. Since in classification task, data tend to cluster based on predicted categories in feature space and $\mathcal{T}$ is distributed across all clusters, it is more reasonable to predict $\mathcal{T}$ using clusters as states. Thus, we define the state space as $(\mathcal{L}^*_c, \mathcal{U}_c)$, where $\mathcal{U}_c$ refers to the $c$-th cluster from the unlabeled data pool $\mathcal{U}$, and $\mathcal{L}^*_c$ is a cluster from the labeled dataset $\mathcal{L}$ defined as:
\begin{equation}
\mathcal{L}^*_c = \argmin_m d(\mathcal{L}_m, \mathcal{U}_c)
\end{equation}
In this equation, $\mathcal{L}_m$ refers to the $m$-th cluster from $\mathcal{L}$, and $\mathcal{L}^*_c$ is the closest labeled cluster to $\mathcal{U}_c$ based on the distance function $d(\cdot, \cdot)$. We use the Wasserstein Distance~\citep{flamary2021pot} in our RL process. For each $(\mathcal{L}, \mathcal{U})$ sample, there are $C$ states, where $C$ is the number of clusters in $\mathcal{U}$. For improved efficiency, we extract stochastic features of clusters as input for the policy, specifically using mean and variance as $[E[\mathcal{L}^*_c], Var[\mathcal{L}^*_c], E[\mathcal{U}_c], Var[\mathcal{U}_c]]$.

\textbf{Action $\mathcal{U}_c^a$.} As data in the TrustSet tend to group together by cluster, we further divide $\mathcal{U}_c$ into $A_c$ data groups, denoted as $\{\mathcal{U}^a_c\}_{a=1}^{A_c}$. Given $(\mathcal{L}^*_c, \mathcal{U}_c)$ as input, the policy selects the top clusters within $\{\mathcal{U}^a_c\}_{a=1}^{A_c}$ that have high potential to be included in the TrustSet. Consequently,  $\{\mathcal{U}_c^a\}_{a=1}^{A_c}$ represents the candidate action space for each state, and the union of the selected actions forms the final selected data subset $\mathcal{S}$.

\textbf{Reward $R$.} Since different $\mathcal{U}^a_c$ contain varying numbers of data points, for a fixed size of $\mathcal{S}$, we need to select a varying number of $\mathcal{U}^a_c$. It is more general to define the reward based on $\mathcal{U}^a_c$ rather than $\mathcal{S}$. We set the reward as the negative Wasserstein distance between $\mathcal{U}^a_c$ and the sub-TrustSet $\mathcal{T}_c=\mathcal{T}\cap\mathcal{U}_c$ as:
\begin{equation}
\label{eq:reward}
R = -d(\mathcal{U}^a_c, \mathcal{T}_c)
\end{equation}
where $\mathcal{U}^a_c$ closer in distribution to $\mathcal{T}_c$ receives better reward.

We follow the DQN method~\citep{mnih2013playing} and illustrate the overall RL process in Figure~\ref{fig:rl}. Given the dataset $\mathcal{L}$ and $\mathcal{U}$, we pass them through the target model $M_{\theta_i}$ to obtain the feature space. To construct the input state for the policy, we cluster the features of $\mathcal{L}$ into $M$ clusters, denoted as $\{\mathcal{L}_m\}^M_{m=1}$ and the features of $\mathcal{U}$ into $C$ clusters as $\{\mathcal{U}_c\}^C_{c=1}$. To generate candidate actions, we further divide $\mathcal{U}_c$ into $A_c$ clusters, denoted as $\{\mathcal{U}^a_c\}_{a=1}^{A_c}$. Meanwhile, we extract the TrustSet for each cluster as $\mathcal{T}_c=\mathcal{T}\cap \mathcal{U}_c$. 

Unlike traditional RL tasks, we consider the future effect in curriculum learning and TrustSet selection, focusing only on the next timestep for policy training. As a result, training the Q-function is equivalent to training the reward function $R_\phi$ as:
\begin{equation}
\label{eq:q_fun}
\Tilde{R} = R_{\phi} (\mathcal{U}_c, \mathcal{L}^*_c, \mathcal{U}^a_c)
\end{equation}
where $\Tilde{R}$ refers to the predicted reward value by $R_{\phi}$ and $\phi$ refers to the parameter of the reward function. And $\phi$ is updated by the Mean Square Error (MSE) loss as:
\begin{equation}
l_{RL} = E_{(\mathcal{U}_c, \mathcal{U}^a_c)}\|R - \Tilde{R}\|
\label{eq:dqn}
\end{equation}
During RL, we calculate the reward for all candidate actions and states to optimize the reward function $R_\phi$. During the active learning process, for each unlabeled cluster $U_c$ in the real unlabeled data pool $U$, we predict the reward for all candidate actions $U^a_c$ and select them in descending order based on the reward score until the fixed size of subset selection is satisfied.


It is worthwhile to note that for each $(\mathcal{L}, \mathcal{U})$ sampled from $L$, $M_\theta$ needs to be retrained on $\mathcal{D}=\mathcal{L}\cup\mathcal{U}$ to extract the TrustSet $T_\mathcal{D}$ as well as $\mathcal{T}$, since the most important data for the model may vary depending on the labeled training set. To avoid the time-consuming process of frequent retraining, we approximate the extraction of $T_\mathcal{D}$ by reusing $M_{\theta_{L_i}}$ which has been trained on $L_i$ in the $i$-th active learning iteration. This is justified by the fact that $\mathcal{D}$ is randomly sampled from $L_i$ and our main objective is to enhance $L_i$ based on the performance of $M_{\theta_{L_i}}$ in the $i$-th active learning iteration. The only requirement is to retrain the policy from scratch for each active learning iteration. In practice, we extract $T_i$ from the entire labeled set $L_i$ and extract $\mathcal{T}$ by taking the intersection between $\mathcal{U}$ and $T_{i}$ as $\mathcal{T}=T_i\bigcap\mathcal{U}$ for efficiency. Please refer to the Appendix for more details of RL training.

\section{Experiment}
\label{sec:exp}


In this section, we evaluate our proposed BRAL-T method on the image classification task and compare our results with previous active learning baselines, following the experimental settings of \citet{zhan2022comparative}. Additionally, we also evaluate BRAL-T on the active fine-tuning task~\citep{xie2023active} and compare it with the current state-of-the-art method, ActiveFT. 
More experiments will be presented in Appendix.

\subsection{Image Classification Results}
\label{sec:expcls}

\textbf{Datasets:} We evaluated BRAL-T on the image classification task across 8 benchmarks, including Cifar10, Cifar100~\citep{krizhevsky2009learning}, Cifar10-imb, EMNIST~\citep{cohen2017emnist}, FashionMNIST~\citep{xiao2017fashion}, BreakHis \citep{spanhol2015dataset}, Pneumonia-MNIST~\citep{kermany2018identifying} and Waterbird~\citep{sagawa2019distributionally, koh2021wilds}. To create the Cifar10-imb dataset, we subsampled the training set of Cifar10 with ratios of 1:2:...:10 for classes 0 through 9.


\textbf{Baselines:} We compared BRAL-T with three baselines, LossPrediction~\citep{yoo2019learning}, WAAL~\citep{shui2020deep} and RandomSample. LossPrediction employs an additional module that predicts the loss for each data point. WAAL adopts min-max loss to better distinguish labeled and unlabeled samples while searching unlabeled batch with higher diversity than labeled samples. According to the experiments in \citet{zhan2022comparative}, among all the methods, LossPrediction and WAAL achieve best results in 6 benchmarks and competitive results in other 2 benchmarks, therefore we select them as our baselines. For RandomSample, we randomly selected a subset from the unlabeled dataset in each active learning iteration. Besides, we visualized accuracy-budget curve on Cifar10, Cifar10-imb, Cifar100 and FashionMNIST benchmarks and compared with LossPrediction, WAAL, VAAL~\citep{sinha2019variational}, BADGE~\citep{ash2019deep}, CoreSet~\citep{zhan2022comparative}, Cluster-Margin~\citep{citovsky2021batch}, BALD~\citep{gal2017deep} and KMeans~\citep{ash2019deep}. To ensure a fair comparison, we used ResNet18~\citep{he2016identity} as the target model. For more experimental details and hyperparameter settings, please refer to Appendix.

\textbf{Evaluation Metrics:} 
For all benchmarks, we report evaluation results using two metrics: \textit{area under the budget curve} (AUBC)~\citep{zhan2021multiple, zhan2021comparative} and \textit{final accuracy} (F-acc). AUBC refers to the area under the accuracy-budget curve. Methods with a higher AUBC score achieve better overall performance across different sizes of the training set. F-acc refers to the final accuracy achieved after the budget $Q$ is exhausted. The experiments for BRAL-T and the baselines were repeated for 3 trials under different random seeds, and the average of the evaluation results are reported.

\begin{table*}[t]
    \centering
    \scriptsize
    \begin{tabular}{l|cc|cc|cc|cc}
        \hline
        \multirow{2}{4em}{\textbf{Methods}} & \multicolumn{2}{c|}{\textbf{FashionMNIST}} & \multicolumn{2}{c|}{\textbf{EMNIST}} & \multicolumn{2}{c|}{\textbf{CIFAR10}} & \multicolumn{2}{c}{\textbf{CIFAR100}} \\
        & \textbf{AUBC} & \textbf{F-acc} & \textbf{AUBC} & \textbf{F-acc} & \textbf{AUBC} & \textbf{F-acc} & \textbf{AUBC} & \textbf{F-acc} \\
        \hline
        \textbf{LossPrediction}  & 0.859 & 0.888 & 0.762 & 0.793 & 0.837 & 0.911 & 0.481 & 0.655 \\
        \textbf{WAAL} & 0.861 & 0.891 & 0.808 & 0.831 & 0.842 & 0.883 & 0.460 & 0.594 \\
        \textbf{RandomSample}  & 0.844 & 0.874 & 0.804 & 0.828 & 0.832 & 0.902 & 0.517 & 0.650 \\
        \textbf{BRAL-T} & \textbf{0.863} & \textbf{0.894} & \textbf{0.813} & \textbf{0.833} & \textbf{0.847} & \textbf{0.916} & \textbf{0.525} & \textbf{0.662} \\
        \hline
        
    \end{tabular}
    
    \begin{tabular}{cc}
    \end{tabular}

    \begin{tabular}{l|cc|cc|cc|cc}
        \hline
        \multirow{2}{4em}{\textbf{Benchmarks}} & \multicolumn{2}{c|}{\textbf{Cifar10-imb}} & \multicolumn{2}{c|}{\textbf{BreakHis}} & \multicolumn{2}{c|}{\textbf{Pneum.MNIST}} & \multicolumn{2}{c}{\textbf{Waterbird}} \\
        & \textbf{AUBC} & \textbf{F-acc} & \textbf{AUBC} & \textbf{F-acc} & \textbf{AUBC} & \textbf{F-acc} & \textbf{AUBC} & \textbf{F-acc} \\
        \hline
        \textbf{LossPrediction} &0.748 & 0.848 & 0.834 & 0.844 & 0.732 & 0.870 & 0.588 & 0.586 \\
        \textbf{WAAL} & 0.752 & 0.799 & 0.836 & 0.855 & 0.640 & 0.870 & 0.525 & 0.506 \\
        \textbf{RandomSample} & 0.710 & 0.810 & 0.834 & 0.832 & 0.706 & 0.652 & 0.586 & 0.502 \\
        \textbf{BRAL-T} & \textbf{0.762} & \textbf{0.851} & \textbf{0.849} & \textbf{0.868} & \textbf{0.738} & \textbf{0.883} & \textbf{0.606} & \textbf{0.618} \\
        \hline
    \end{tabular}


    \caption{Experiment results of image classification task on 8 benchmarks.
    }
    \label{tab:result}
\end{table*}

\begin{figure*}
    \centering
    \includegraphics[width=.9\textwidth]{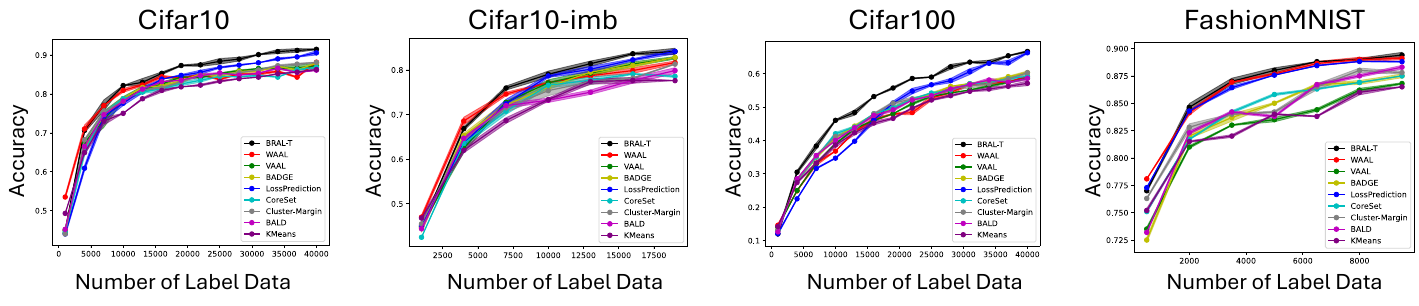}
    \caption{Visualization of experiment results on Cifar10, Cifar10-imb, Cifar100 and FashionMNIST. 
    }
    \label{fig:vis_result}
\end{figure*}

\textbf{Experiment Results:} The experimental results on 8 benchmarks are presented in Table~\ref{tab:result}. BRAL-T significantly outperforms RandomSample under all benchmarks. Compared with WAAL and LossPrediction, BRAL-T achieves better AUBC as well as F-acc on all benchmarks. In Figure~\ref{fig:vis_result}, we visualize the accuracy-budget curves of BRAL-T and baselines on 4 benchmarks. BRAL-T consistently achieves higher accuracy throughout the entire active learning process for Cifar100 and FashionMNIST. In early active learning iterations of Cifar10 and Cifar10-imb, BRAL-T has a bit worse accuracy compared with WAAL, the reasons of which could be attributed to WAAL's emphasis on diversity. However, without adequate consideration for uncertainty cause performance diminishing of WAAL when the size of labeled dataset increases. As comparison, LossPrediction focuses solely on uncertain data with high predicted loss, neglecting the diversity of the selected subset, which results in bad performance in early stage. 

\subsection{Active Learning on More Long-tail Datasets}
\label{sec:lt_exp}



\textbf{Dataset.} Besides the aforementioned benchmarks, in this section, we focus on long-tail datasets, including CIFAR10-LT and CIFAR100-LT. Both datasets are subsampled from CIFAR datasets and the number of samples within each classes decreases exponentially with factor within 10 and 100. Specifically, we consider 10 and 20 in our experiments. The test images of CIFAR10-LT and CIFAR100-LT are the same as those in CIFAR10 and CIFAR100 datasets respectively. Both the two benchmarks are open-source and can be accessed through huggingface \textit{tomas-gajarsky/cifar10-lt} and \textit{tomas-gajarsky/cifar10-lt}. Please refer to Appendix for more details.

\begin{table}[t]
    \centering
    \scriptsize
    \begin{tabular}{l|cc|cc|cc|cc}
        \hline
        \multirow{2}{4em}{\textbf{Methods}} & \multicolumn{2}{c|}{\textbf{Cifar10-LT-r10}} & \multicolumn{2}{c|}{\textbf{Cifar10-LT-r20}} & \multicolumn{2}{c|}{\textbf{Cifar100-LT-r10}} & \multicolumn{2}{c}{\textbf{Cifar100-LT-r20}} \\
        & \textbf{AUBC} & \textbf{F-acc} & \textbf{AUBC} & \textbf{F-acc} & \textbf{AUBC} & \textbf{F-acc} & \textbf{AUBC} & \textbf{F-acc} \\
        \hline
        \textbf{TiDAL}& 0.510 & 0.619 & 0.435 & 0.583 & 0.398 & 0.450 & 0.365 & 0.420 \\
        \textbf{SIMILAR}  & 0.507 & 0.654 & 0.431 & 0.566 & 0.405 & 0.514 & 0.366 & 0.460 \\
        \textbf{LossPrediction}  & 0.478 & 0.679 & 0.413 & 0.585 & 0.424 & 0.516 & 0.380 & 0.460 \\
        \textbf{WAAL} & 0.510 & 0.623 & 0.435 & 0.589 & 0.381 & 0.458 & 0.342 & 0.412 \\
        \textbf{RandomSample}  & 0.476 & 0.686 & 0.397 & 0.587 & 0.382 & 0.463 & 0.343 & 0.410 \\
        \textbf{BRAL-T} & \textbf{0.512} & \textbf{0.686} & \textbf{0.440} & \textbf{0.614} & \textbf{0.432} & \textbf{0.517} & \textbf{0.384} & \textbf{0.462} \\
        \hline
    \end{tabular}

    \caption{Experiment results on Cifar10-LT and Cifar100-LT datasets. }
    \label{tab:lt_data}
\end{table}


\textbf{Experiment Results.} Besides LossPrediction, WAAL and RandomSample, we also compare BRAL-T with SIMILAR~\citep{kothawade2021similar} and TiDAL~\citep{kye2023tidal} which are designed for active learning in imbalanced dataset. As shown in Table~\ref{tab:lt_data},  BRAL-T achieves the best AUBC as well as F-acc results. As pseudo label is not reliable especially when target model might be overconfident in long-tail dataset, BRAL-T selects informative data with ground-truth label to construct TrustSet which is more reliable to reflect whether target model has sufficiently learnt from related samples. As a result, BRAL-T always performs better than SIMILAR. Moreover, compared with LossPrediction and WAAL, we encourage TrustSet to be balanced which releases the category bias problem in long-tail distribution and contributes to the success of BRAL-T.

\subsection{Active Learning for Finetuning Results}
\label{sec:activeft}


\textbf{Experiment Setting:} 
We adhere to the settings of \citet{xie2023active} and use Deit-Small~\citep{touvron2021training}, pretrained with the DINO~\citep{caron2021emerging} framework on ImageNet-1k, as the target model. We chose two datasets for fine-tuning: Cifar10-imb and TinyImageNet~\citep{le2015tiny}, resizing all images to $224\times224$. For more implementation details, we utilize ActiveFT~\citep{xie2023active} to select an $1\%$ subset as the initial labeled dataset and select an additional $1\%$ of data for each active learning iteration. The pretrained model is fine-tuned using the SGD optimizer for 1000 epochs with a batch size of 512. Cosine learning rate decay is applied during the fine-tuning phase of each active learning iteration.

\begin{table}
    \begin{minipage}{.5\linewidth}
    \centering
    \scriptsize
    \setlength\tabcolsep{1.pt}
    \begin{tabular}{l|cc|cc}
    \hline
       \multirow{2}{*}{\textbf{Methods}} & \multicolumn{2}{c|}{\textbf{Cifar10-imb}} & \multicolumn{2}{c}{\textbf{TinyImageNet}} \\
         & $2\%$ & $3\%$ & $2\%$ & $3\%$ \\\hline
         \textbf{RandomSample} & 0.841 & 0.856 & 0.213 & 0.348 \\
         \textbf{ActiveFT} & 0.838 & 0.851 & 0.289 & 0.359 \\
         \textbf{BRAL-T} & \textbf{0.852} & \textbf{0.865} & \textbf{0.300} & \textbf{0.392}\\
    \hline
    \end{tabular}
    \caption{Results of Active Finetuning task.}
    \label{tab:activeft}
    \end{minipage}%
    \begin{minipage}{.5\linewidth}
    \centering
    \scriptsize
    \setlength\tabcolsep{.9pt}
    \setlength\tabcolsep{2pt}
    \begin{tabular}{l|cc|cc}
    \hline
        \multirow{2}{4em}{\textbf{Baseline}} & \multicolumn{2}{c|}{\textbf{Cifar10}} & \multicolumn{2}{c}{\textbf{Cifar100}} \\
         & \textbf{AUBC} & \textbf{F-acc} & \textbf{AUBC} & \textbf{F-acc} \\\hline
        \textbf{PseudoScore} & 0.842 & 0.908 & 0.486 & 0.661  \\
        \textbf{BRAL-DiffSet} & 0.843 & 0.909 & 0.521 & 0.652  \\
        \textbf{BRAL-T w/o CL} & 0.845 & 0.906 & 0.522 & 0.662 \\\hline
        \textbf{RandomSample} & 0.832 & 0.902 & 0.517 & 0.650 \\
        \textbf{BRAL-T} & \textbf{0.847} & \textbf{0.916} & \textbf{0.525} & \textbf{0.662}\\
        \hline
    \end{tabular}
    \caption{Ablation study result.}
    \label{tab:ablation}
    \end{minipage}
\end{table}

\textbf{Experiment Results:} All experiments were repeated across 3 trials, and the average results are reported in Table~\ref{tab:activeft}. BRAL-T significantly outperforms the other baselines. ActiveFT could suffer from the long-tail distribution of unlabeled data pool while TrustSet are defined to be class-balanced.

\subsection{Ablation Study}
\label{sec:abl}

\textbf{Experiment Setting:} To further evaluate BRAL-T, we conducted ablation studies to demonstrate the benefits of the proposed modules by considering three baselines. For \textbf{PseudoScore}, instead of training an RL policy, we assign pseudo-labels to the unlabeled data pool based on the category with the highest logit score. Data subset with top EL2N score will be selected during active learning. For \textbf{BRAL-DiffSet}, to show the effectiveness of TrustSet, We select the second-best data group instead of the best as TrustSet. For \textbf{BRAL-T w/o CL}, we remove curriculum learning and directly use the cross-entropy loss function to calculate the EL2N score.


\textbf{Experiment Results:} Table~\ref{tab:ablation} displays the results for the Cifar10 and Cifar100 datasets. BRAL-T surpasses PseudoScore in both AUBC and F-acc as pseudo-labels are often inaccurate, especially when classifier has low performance. In contrast, selecting the TrustSet based on the labeled dataset is more reliable. Compared to BRAL-DiffSet, BRAL-T also achieves better AUBC and F-acc scores, empirically proving the correlation between EL2N score and model accuracy. Moreover, curriculum learning also plays a crucial role in the success of BRAL-T, which aids in selecting easy examples and enhances the performance of the target model in the initial stages.

\section{Conclusion}

In summary, our RL-based Active Learning framework, BRAL-T, leverages TrustSet to more accurately evaluate distribution of labeled datasets and employs an RL policy to learn from TrustSet. BRAL-T benchmarked against 8 baselines across 8 image classification tasks, shows superior AUBC and F-acc performance. Moreover, in CIFAR-LT benchmarks, BRAL-T outperforms baselines to handle long-tail dataset. Additionally, its application in active fine-tuning tasks reveals new state-of-the-art results.

\bibliography{iclr2026_conference}
\bibliographystyle{iclr2026_conference}

\appendix

\section{The Usage of Large Language Model (LLM)}

LLM is only leveraged to polish the writing and correct wrong expressions in the paper. We provide each paragraph of the paper draft to ChatGPT separately and ask for revision. The core ideas, contributions of the paper and technique details are originated by the author without the LLM engagement.

\section{Method Details}
\label{sec:ap_detail}

BRAL-T comprise two iterative processes: active learning process and reinforcement learning process. Algorithm~\ref{alg:overall} shows the pseudocode of overall framework. We randomly sampled initial labeled dataset and initialize parameters of target model and reward network in lines 3-5. During the $i$-th active learning process (lines 7-9), we trained target model $M_{\theta_i}$ with $i$-th labeled dataset $L_i$ from scratch and extract TrustSet $T_i$ from $L_i$, details of which is depicted in Section~\ref{sec:method_trustset}. During the reinforcement learning (RL) process (lines 11-17), we followed DQN~\citep{mnih2013playing} and initialized replay buffer $\mathbb{B}$ to be empty. For each RL iteration, we sample labeled set and unlabeled set from $L_i$ and store state set $\{\mathcal{L}^*_c, \mathcal{U}_c, \mathcal{U}^a_c, T_c\}$ into replay buffer (detailed in algorithm~\ref{alg:ext}). To train reward function $R_{\phi_i}$, a data batch is sampled from $\mathbb{B}$ and parameters of $R_{\phi_i}$ is updated based on Eq~\ref{eq:dqn} (detailed in algorithm~\ref{alg:rl}). After the two processes, we sampled a new dataset for oracle to annotate and updated $L_i$ and $U_i$.

\begin{algorithm}[h]
   \caption{BRAL-T}
   \label{alg:overall}
\begin{algorithmic}[1]
   \STATE {\bfseries Input:} Dataset $D$
   \STATE {\bfseries Output:} Target Model $M_\theta$
   \STATE Random sample $L_0$ from $D$ and annotated by oracle;
   \STATE Set $U_0 := D\setminus L_0$;
   \STATE Initialize $M_{\theta_0}$, $R_{\phi_0}$;\\
   \FOR{$i=0$ {\bfseries to} $N-1$}
       \STATE // Active Learning Process
       \STATE Train $M_{\theta_i}$ with $L_i$ from scratch;
       \STATE Extract TrustSet $T_i$ from $L_i$;
       \STATE
       \STATE // Reinforcement Learning Process
       \STATE Initialize Replay Buffer $\mathbb{B}$;
       \FOR{$j=0$ {\bfseries to} $K$}
            \STATE Sample $\mathcal{L}$ and $\mathcal{U}$ from $L_i$;
            \STATE Extract set $\{\mathcal{L}^*_c, \mathcal{U}_c, \mathcal{U}^a_c, \mathcal{T}_c\}=\mathbb{E}(\mathcal{L},\mathcal{U},T_i)$ and store into $\mathbb{B}$; (Algorithm~\ref{alg:ext})
            \STATE Sample data from $\mathbb{B}$ and train $R_{\phi_i}$ as Eq~\ref{eq:dqn}. (Algorithm~\ref{alg:rl})
       \ENDFOR
       \STATE
       \STATE // Sample New DataSet
       \STATE Sample $S_i:=\pi(R_{\phi_i}, L_i, U_i)$;
       \STATE Update $L_{i+1}:=L_i\cup S_i$ and $U_{i+1}:=U_i\setminus S_i$;
       
   \ENDFOR

\end{algorithmic}
\end{algorithm}

In algorithm~\ref{alg:ext}, we show the pseudocode of data extraction for RL (line 15 of algorithm~\ref{alg:overall}). As illustrated in Section~\ref{sec:method_rl}, we clustered labeled set into $\{L_m\}^M_{m=1}$ and unlabeled set into $\{U_c\}^C_{c=1}$ to formulate state space of RL. For each unlabeled subset, we further cluster $U_c$ into $\{U^a_c\}^{A_c}_{a=1}$ to formulate action space of RL and extract Trustset $T_c$ for each $U_c$. All pairs of $\{L^*_c, U_c, U_c^a, T_c\}$ are stored and return as extraction results.
\begin{algorithm}[]
    \caption{Data Extraction For Reinforcement Learning}
    \label{alg:ext}
    \begin{algorithmic}[1]
        \STATE {\bfseries Input:} LabeledSet $\mathcal{L}$, UnlabeledSet $\mathcal{U}$, TrustSet $T$
        \STATE {\bfseries Output:} Data list $Out$
        \STATE Initialize output list $Out:=[]$;
        \STATE Cluster $\mathcal{L}$ into $\{\mathcal{L}_m\}^M_{m=1}$;
        \STATE Cluster $\mathcal{U}$ into $\{\mathcal{U}_c\}_{c=1}^C$
        \FOR{\textbf{each} $U_c$}
        \STATE Extract $\mathcal{T}_c:=T_i\cap \mathcal{U}_c$
        \STATE Cluster $\mathcal{U}_c$ into $\{\mathcal{U}_c^a\}_{a=1}^{A_c}$;
        \STATE Calculate $\mathcal{L}^*_c:= \argmin_m d(\mathcal{L}_m, \mathcal{U}_c)$;
        \STATE Store each $\{\mathcal{L}^*_c, \mathcal{U}_c, \mathcal{U}_c^a, \mathcal{T}_c\}$ into $Out$;
        \ENDFOR
        \STATE Return $Out$;
    \end{algorithmic}
\end{algorithm}

In algorithm~\ref{alg:rl}, we show the pseudocode of reinforcement learning to train data selection policy (line 16 of algorithm~\ref{alg:overall}). For each gradient step, we sample state and action data from replay buffer $\mathbb{B}$ and extract vector input $S$ and $A$ (line 4-6). Then based on Eq.~\ref{eq:reward}, we calculate the reward for each (state, action) pair as negative distance between data subset and TrustSet (line 7). And based on Eq.~\ref{eq:q_fun}, we predict reward with current reward function $R_\phi$ (line 8). Finally, we calculate mean square error (MSE) loss between predicted reward $r$ and ground truth reward $R$ and update reward function with gradient descent (line 9).

\begin{algorithm}[]
    \caption{Training of Reinforcement Learning}
    \label{alg:rl}
    \begin{algorithmic}[1]
        \STATE {\bfseries Input:} Replay Buffer $\mathbb{B}$, Reward Function $R_\phi$.
        \STATE {\bfseries Output:} Update Reward Function $R_\phi$
        \FOR{Each Gradient Step}
        \STATE Sample data batch from $\mathbb{B}$ as $\{L^*_c, U_c, U_c^a, T_c\}^B$.
        \STATE Extract state vector input as: $S=[E[L^*_c], Var[L^*_c], E[U_c], Var[U_c]]$.
        \STATE Extract action input as: $A=[E[U_c^a], Var[U_c^a]]$.
        \STATE Calculate reward for each state action pair as Eq.~\ref{eq:reward}: $R=-d(A, T_c)$.
        \STATE Predicate reward with $R_\phi$ as Eq.~\ref{eq:q_fun}: $r=R_\phi(S, A)$.
        \STATE Calculate Loss $L=\text{MSE}(R, r)$ and update $R_\phi$ with gradient descent.
        \ENDFOR

        \STATE Return $R_\phi$;
    \end{algorithmic}
\end{algorithm}

\section{Experiment Details}
\label{sec:ap_exp}

In this section, we introduce more experiment details of Section~\ref{sec:exp}, including architecture of target model we used for image classification and hyperparameter settings of experiments.

\begin{table}[h]
    \centering
    \setlength\tabcolsep{2.5pt}
    \begin{tabular}{lccccccc}
        \hline
        Benchmarks & $|L_0|$ & $|U_0|$ & $Q$ & $b$ & $\#e$ & $C$ \\
        \hline
        FashionMNIST & 500 & 59,500 & 10,000 & 250 &  40 & 10 \\
        EMNIST & 1,000 & 696,932 & 50,000 & 500 & 40 & 62 \\
        CIFAR10 & 1,000 & 49,000 & 40,000 & 500 & 50 & 10 \\
        CIFAR100 & 1,000 & 49,000 & 40,000 & 500 & 60 & 100 \\
        CIFAR10-imb & 1,000 & 27,239 & 20,000 & 500 & 50 & 10\\
        CIFAR10-LT & 1,000 & - & $|D|$ & 100 & 50 & 10\\
        CIFAR100-LT & 4,000 & - & $|D|$ & 500 & 60 & 100\\
        BreakHis & 100 & 5,436 & 5,000 & 100 & 30 & 2 \\
        PneumoniaMNIST & 100 & 5,132 & 5,000 & 100 & 30 & 2 \\
        Waterbird & 100 & 4,695 & 4,000 & 100 & 30 & 2 \\
        \hline
    \end{tabular}
    \caption{Setting of benchmarks. Where $|L_0|$ refers to size of initial labeled set, $|U_0|$ refers to size of initial unlabeled data pool, $Q$ refers to budget, $b$ refers to batch size for target model training, $\#e$ refers to number of epoch for target model training and $C$ refers to number of clusters from unlabeled data pool. For all the benchmarks, the number of clusters $M$ from labeled dataset is set to be the same as $C$ and number of candidate action for $U_c$ is set to be 5.}
    \label{tab:benchmark}
\end{table}

\textbf{DataSets.} We evaluated BRAL-T on the image classification task across 5 benchmarks, including Cifar10, Cifar100~\citep{krizhevsky2009learning}, Cifar10-imb, EMNIST~\citep{cohen2017emnist}, and FashionMNIST~\citep{xiao2017fashion}. To create the Cifar10-imb dataset, we followed the settings of \citet{zhan2022comparative} and subsampled the training set with ratios of 1:2:...:10 for classes 0 through 9. We also evaluated our framework on medical imaging analysis tasks across 2 benchmarks, including Breast cancer Histopathological Image Classification (BreakHis)~\citep{spanhol2015dataset} and Chest X-Ray Pneumonia classification (Pneumonia-MNIST)~\citep{kermany2018identifying}. Additionally, we assessed our framework on an object recognition dataset with correlated backgrounds (Waterbird)~\citep{sagawa2019distributionally,koh2021wilds}, which contains waterbird and landbird classes manually mixed with water and land backgrounds. To further evaluate BRAL-T on long-tail datasets, we also consider CIFAR10-LT and CIFAR100-LT where the number of samples within each classes decreases exponentially with factor to be 10, 20 or 50. 

The detail setting for each benchmark are shown in Table~\ref{tab:benchmark}, including initial data size of labeled dataset $|L_0|$ and unlabeled dataset $|U_0|$, final budget $Q$ of labeled dataset, batch size $b$ for data subset selection in each active learning iteration, training epoch $\#e$ for target model training, and category number $C$ for dataset.

\textbf{Model Details.} Following the setting of \citet{zhan2022comparative}, we use Resnet18~\citep{he2016identity} as the target model for image classification tasks. For Cifar10, Cifar10-imb, Cifar100 and PneumoniaMNIST, we replaced the kernal size of first convolutional layer to be $3\times 3$ and stride to be 1 in order to handle image with smaller size. For grayscale images such as FashionMNIST and EMNIST datasets, we add an additional convolutional layer before the first layer of Resnet with $1\times 1$ kernal to increase the channel number of images to be 3. Furthermore, we trained the target models of all baselines for the same number of epochs, as shown in Table~\ref{tab:benchmark}. For LossPrediction, the target model is trained with both classification loss and loss prediction loss for the first 20 epochs. After 20 epochs, only the gradient from the classification loss is back-propagated through the target model.

\textbf{Model and Hyperparameters Setting:} We constructed the reward function $R_\phi$ using a fully connected network comprising 2 hidden layers, each with 512 units, and use the ReLU activation function. SGD was employed as the optimizer for $R_\phi$, with the learning rate set at 0.01. For hyperparameters of curriculum learning, we follow the setting of SuperLoss~(\citep{castells2020superloss}) and set $\tau=\log |K|$ where $|K|$ is the category number. Additionally, we set the value of $\lambda$ to be 0.25 for EMNIST, CIFAR100 and TinyImageNet datasets and 1.0 for the others. During active learning, We train target model with SGD optimizer for PneumoniaMNIST and Waterbird benchmarks and Adam optimizer for other datasets. 

After each active learning iteration, we sampled 30 pairs of $\mathcal{L}$ and $\mathcal{U}$ from the existing labeled set $L$ to train the policy, setting the batch size to 100 pairs of state, action, and reward. Following each sampling, we trained $R_\phi$ for 20 iterations, resulting in a total of 600 iterations for the entire RL training process. As shown in Table~\ref{tab:benchmark}, the number of clusters $C$ for unlabeled set and $M$ for labeled set are set to be the same as category number for related benchmark. And the number of candidate action $A_c$ for each unlabeled cluster $U_c$ is set to be 5 during the experiment.

\section{More Experiment Results}
\label{sec:ap_exp_res}

In this section, we introduce more experiments and results. First of all, we show the confidence interval results for Table~\ref{tab:result} over 8 benchmarks in \ref{sec:ci_res}. Then we evaluate BRAL-T by calculating penalty matrix in \ref{sec:penalty}. Moreover, to show the efficiency of BRAL-T, we compare time overhead between BRAL-T and baselines in \ref{sec:timecost}. Finally, in \ref{sec:moreablation}, we show more ablation studies of BRAL-T.


\subsection{Confidence Intervals of Results in Image Classification Tasks.}
\label{sec:ci_res}

Besides representing average value of AUBC and F-acc of BRAL-T and baselines on image classifcation benchmarks, Table~\ref{tab:ci_result} shows the confidence interval of experiment results. In general, BRAL-T results are stable and robust over different experiment trials.

\begin{table*}[h]
    \centering
    \setlength\tabcolsep{.9pt}
    \begin{tabular}{l|cc|cc|cc|cc}
        \hline
        \multirow{2}{4em}{\textbf{Methods}} & \multicolumn{2}{c|}{\textbf{FashionMNIST}} & \multicolumn{2}{c|}{\textbf{EMNIST}} & \multicolumn{2}{c|}{\textbf{CIFAR10}} & \multicolumn{2}{c}{\textbf{CIFAR100}} \\
        & \textbf{AUBC} & \textbf{F-acc} & \textbf{AUBC} & \textbf{F-acc} & \textbf{AUBC} & \textbf{F-acc} & \textbf{AUBC} & \textbf{F-acc} \\
        \hline
        \textbf{LossPrediction}  & $\pm$0.002 & $\pm$0.038 & $\pm$ 0.016 & $\pm$ 0.022 & $\pm$0.006 & $\pm$ 0.012 & $\pm$0.019 & $\pm$0.012 \\
        \textbf{WAAL} & $\pm$0.002 & $\pm$0.015 & $\pm$ 0.012 & $\pm$0.015 & $\pm$ 0.006 & $\pm$0.009 & $\pm$0.006 & $\pm$0.011 \\
        \textbf{RandomSample}  & $\pm$0.001 & $\pm$0.009 & $\pm$ 0.004 & $\pm$0.007 & $\pm$0.003 & $\pm$ 0.011 & $\pm$0.003 & $\pm$0.008 \\
        \textbf{BRAL-T} & $\pm$0.001 & $\pm$0.008 & $\pm$0.005 & $\pm$0.014 & $\pm$0.003 & $\pm$0.006 & $\pm$0.004 & $\pm$0.009 \\
        \hline
        
    \end{tabular}
    
    \begin{tabular}{cc}
    \end{tabular}

    \begin{tabular}{l|cc|cc|cc|cc}
        \hline
        \multirow{2}{4em}{\textbf{Benchmarks}} & \multicolumn{2}{c|}{\textbf{Cifar10-imb}} & \multicolumn{2}{c|}{\textbf{BreakHis}} & \multicolumn{2}{c|}{\textbf{Pneum.MNIST}} & \multicolumn{2}{c}{\textbf{Waterbird}} \\
        & \textbf{AUBC} & \textbf{F-acc} & \textbf{AUBC} & \textbf{F-acc} & \textbf{AUBC} & \textbf{F-acc} & \textbf{AUBC} & \textbf{F-acc} \\
        \hline
        \textbf{LossPrediction} &$\pm$0.011 & $\pm$0.017 & $\pm$0.026 & $\pm$0.037 & $\pm$0.023 & $\pm$0.038 & $\pm$0.014 & $\pm$0.097 \\
        \textbf{WAAL} & $\pm$0.008 & $\pm$0.013 & $\pm$0.016 & $\pm$0.042 & $\pm$0.018 & $\pm$0.021 & $\pm$0.011 & $\pm$0.078 \\
        \textbf{RandomSample} & $\pm$ 0.013 & $\pm$ 0.019 & $\pm$0.015 & $\pm$0.050 & $\pm$0.001 & $\pm$0.009 & $\pm$0.005 & $\pm$0.059 \\
        \textbf{BRAL-T} & $\pm$ 0.012 & $\pm$ 0.008 & $\pm$0.017 & $\pm$0.037 & $\pm$0.012 & $\pm$0.013 & $\pm$0.007 & $\pm$0.024 \\
        \hline
    \end{tabular}

    \caption{Confidence Interval of Experiment results of image classification task.}
    \label{tab:ci_result}
\end{table*}

\subsection{Time Overhead Comparison}
\label{sec:timecost}

\begin{figure}[]
\centering
\includegraphics[width=.8\linewidth]{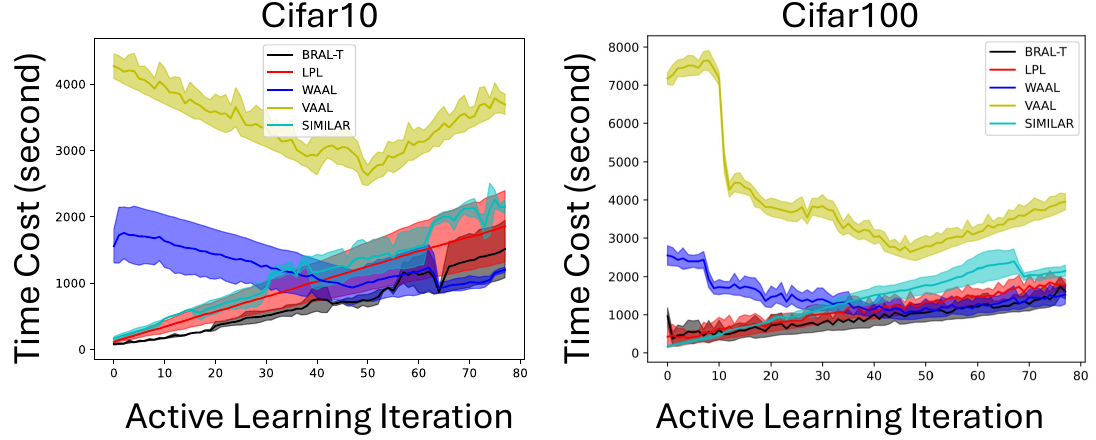}
\caption{Time Cost of BRAL-T and baselines.}
\label{fig:timecost}
\end{figure}

To evaluate the efficiency of BRAL-T, we compare the time overhead with LossPrediction, WAAL, VAAL and SIMILAR on Cifar10 and Cifar100 datasets.  All experiments were conducted using a single Quadro RTS 6000 GPU core with CUDA Version 11.4. Figure~\ref{fig:timecost} shows the time cost results along with active learning iteration. 

The time cost associated with BRAL-T increases with each active learning iteration as the labeled set expands and more samples are clustered during the reinforcement learning process. However, compared to other baselines, BRAL-T consistently demonstrates efficiency, maintaining a competitive edge in terms of computational resource utilization.

\subsection{Pairwise Comparison}
\label{sec:penalty}

We further compare BRAL-T with VAAL~\citep{sinha2019variational}, SAAL~\citep{kim2023saal} and BAIT~\citep{ash2021gone} on Cifar10, Cifar10-imb and FashionMNIST datasets by pairwise penalty matrix following \citet{ash2021gone}. For each benchmark, we collect accuracy results achieved by all baselines. For pairwise comparison between the method for $i$th row ($r_i$) and the method in $j$th column ($c_j$), we add a score to element $e_{ij}$ whenever $r_i$ achieves better accuracy result in one budget of data subset for a benchmark, which means the better $r_i$ performs compared with $c_j$, the higher score $e_{ij}$ will be.
\begin{figure}[h]
    \centering
    \includegraphics[width=.4\textwidth]{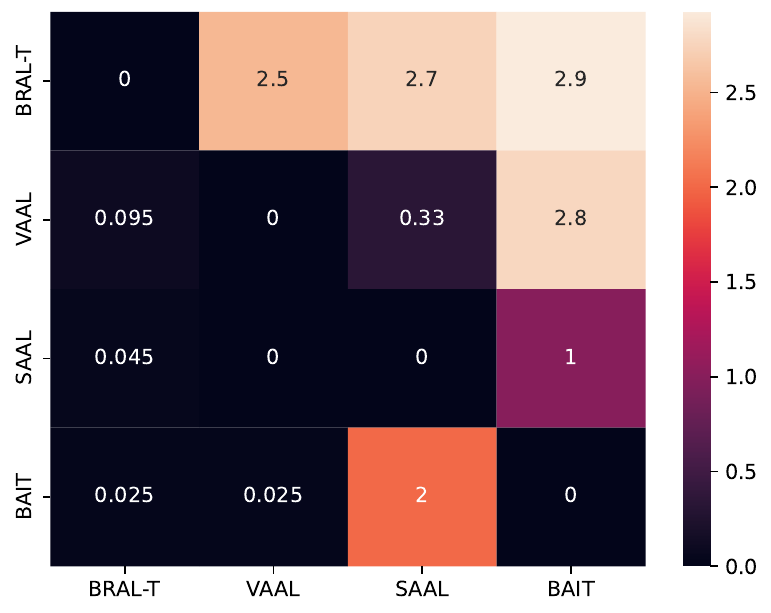}
    \caption{Pairwise Comparison of BRAL-T, VAAL, SAAL and BAIT.}
    \label{fig:penalty}
\end{figure}

Figure~\ref{fig:penalty} represents the pairwise comparison results. Compared with all baselines, BRAL-T achieves highest value in $e_{\text{BRAL-T}, \cdot}$ and lowest value in $e_{\cdot, \text{BRAL-T}}$.





\textbf{Agent Reuse.} A potential way to further improve the efficiency of BRAL-T is reusing RL agent for all active learning iterations. However, considering the distribution shift of labeled dataset, distribution of TrustSet will also shift during active learning. For this reason, we apply two RL agents during active learning, one of which is trained in the first active learning step and remains unchanged for early active learning iterations; the other one of which is maintained for the rest iterations. Specifically for CIFAR10-imb dataset, we use the first agent for the first 20 iterations and the second agent for the rest 20 iterations. The result is shown in Table~\ref{tab:two_agent} below:
\begin{table}[h]
    \centering
    \begin{tabular}{l|cc}
        Method & AUBC & F-Acc \\\hline
        LossPrediction & 0.748 & 0.848 \\
        WAAL & 0.752 & 0.799 \\
        RandomSample & 0.710 & 0.810 \\\hline
        BRAL-T & \textbf{0.762} & \textbf{0.851} \\
        BRAL-T (two agents) & 0.755 & 0.837
    \end{tabular}
    \caption{BRAL-T reusing two RL agents.}
    \label{tab:two_agent}
\end{table}
where BRAL-T with agent reusing surprisingly achieves better AUBC results compares with other baselines. With a more careful separation of active learning stages and RL agents, we believe the performance could be further improved. 

\subsection{More Ablation Study}
\label{sec:moreablation}

To evaluate the robustness of BRAL-T, we run BRAL-T on Cifar10-imb dataset under different qualifies of initial labeled dataset. Moreover, we show the performance of BRAL-T with different candidate action numbers. To evaluate the quality of RL approximation, we apply ground truth labels for TrustSet selection and compare the accuracy results with BRAL-T.

\textbf{Quality Effect of Initial Labeled Set.} We explored the impact of the initial labeled set's quality by applying three different sampling methods to construct the initial labeled set from the Cifar10-imb dataset:
\begin{itemize}
    \item \textbf{Random Sample}: We randomly sample data from the unlabeled pool to form the initial labeled set which maintains a similar category distribution with unlabeled pool.
    \item \textbf{Twisted Main}: We sort the 10 categories by the number of data samples first and then select 50 samples from 5 rare classes and 950 samples randomly from the other 5 main classes.
    \item \textbf{Twisted Rare}: Similar to Twisted Main, we randomly select 50 samples from 5 main classes and 950 samples from the other 5 rare classes.
\end{itemize}


\begin{figure}
    \centering
    \includegraphics[width=.5\textwidth]{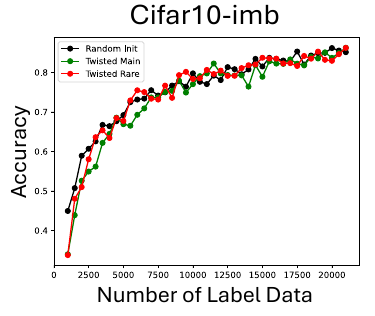}
    \captionof{figure}{Accuracy-budget curve of Different Initial Labeled Set Quality.}
    \label{fig:init_q}
\end{figure}

\begin{table}
    \centering
    \begin{tabular}{c|cc}
        \textbf{Initial Method} & \textbf{AUBC} & \textbf{F-Acc} \\\hline
        \textbf{Random} & 0.762 & 0.851 \\
        \textbf{Twisted Main} & 0.750 & 0.855 \\
        \textbf{Twisted Rare} & 0.756 & 0.855
    \end{tabular}
    \captionof{table}{Experiment Result of Different Initial Labeled Set Quality.}
    \label{tab:init_q}
\end{table}

The results, depicted in the Figure~\ref{fig:init_q}, indicate that BRAL-T's performance varies with the quality of the initial labeled set, particularly when labeled data is scarce. However, as the size of the labeled dataset increases, the accuracy differences become negligible, demonstrating BRAL-T's robustness to the initial set's composition. Despite the initial set's quality impacting BRAL-T's performance, in Table~\ref{tab:init_q}, the AUBC results in the twisted cases are competitive with the results of the WAAL baseline in Table-2, and all achieve better F-Acc compared with other baselines.

\begin{figure}
    \centering
    \includegraphics[width=0.7\textwidth]{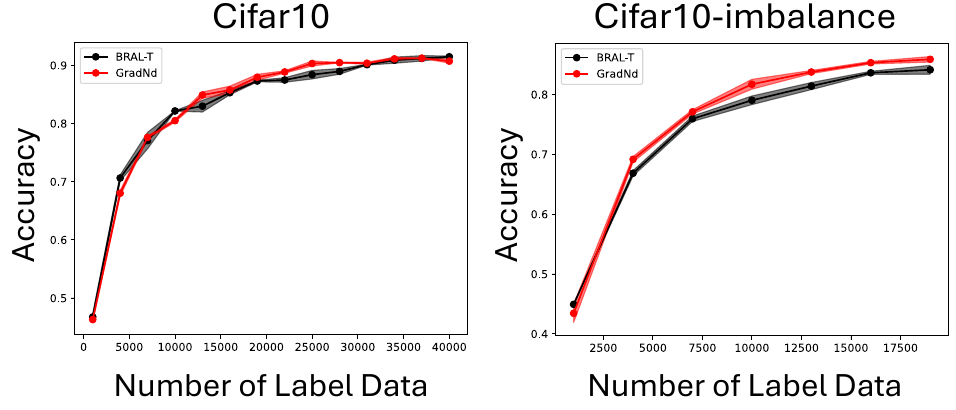}
    \caption{Comparison between BRAL-T with RL policy and Ground Truth Labels.}
    \label{fig:gradnd_res}
\end{figure}

\textbf{Ablation Study on Different Action Numbers.} In the reinforcement learning process, we set number of candidate action to be 5 in Section~\ref{sec:exp}. To evaluate the impact of varying action space sizes, we conducted an ablation study on the Cifar10-imb dataset, comparing BRAL-T's performance across different numbers of actions: 5, 10, 50, and 100. The results are shown in Table~\ref{tab:act_num}

\begin{table}[h]
    \centering
    \begin{tabular}{l|cc}
        \textbf{\# Actions} & \textbf{AUBC} & \textbf{F-Acc} \\\hline
        5 & 0.762 & 0.851 \\
        10 & 0.763 & 0.853 \\
        50 & 0.755 & 0.851 \\
        100 & 0.758 & 0.854 
    \end{tabular}
    \caption{Ablation Study on Different Candidate Action Number.}
    \label{tab:act_num}
\end{table}

Under all different setting of action numbers, BRAL-T achieves best AUBC and F-Acc results compared with baselines in Table~\ref{tab:result}. Setting a large number of actions will increase the complexity of policy training. As we keep the policy architecture to be the same and simple for time efficiency, in some active learning iteration policy might not be trained well with large action number which lead to a small drop of AUBC score. But in general, our method is robust to action number. The reason we choose 5 in the experiment is mainly for the consideration of time efficiency. 

\textbf{Ablation Study on Different Setting of $\lambda$.} During the TrustSet extraction, we introduce curriculum learning where $\lambda$ is introduced to control the effect of SuperLoss. We study the impact of $\lambda$ on the CIFAR10-imb dataset for further sensitivity analysis, the result is shown in table~\ref{tab:lambda}.
\begin{table}[]
    \centering
    \begin{tabular}{c|cc}
        \textbf{$\lambda$} & \textbf{AUBC} & \textbf{F-Acc} \\\hline
        0.25 & \textbf{0.762} & \textbf{0.851} \\
        1.00 & 0.752 & 0.842 \\
        2.00 & 0.749 & 0.830
    \end{tabular}
    \caption{Impact of $\lambda$ value on CIFAR10-imb dataset.}
    \label{tab:lambda}
\end{table}

Increasing the value of $\lambda$ reduces the influence of SuperLoss on the task loss. In an imbalanced dataset, data samples are limited, especially in rare classes. Focusing on difficult data during the early stages of active learning can significantly increase the difficulty of model training. As a result, increasing $\lambda$ leads to a reduction in AUBC and F-Acc for BRAL-T, highlighting the importance of incorporating curriculum learning into the active learning process. However, overall, the AUBC and F-Acc values remain competitive with the baselines presented in Table~\ref{tab:result} of the paper.

\textbf{Compare between RL and Ground Truth Labels.} Although label information of unlabeled data pool is not available during active learning, in order to evaluate the approximation performance of RL policy, for baseline GradNd we assume ground truth label of unlabeled data pool is available when calculating the GradNd score of data samples and we pick class-balanced data with top GradNd score for each active learning iteration. We compare BRAL-T with GradNd on Cifar10 and Cifar10-imb datasets and shows the accuracy-budget results in Figure~\ref{fig:gradnd_res}.

In Cifar10 dataset, BRAL-T achieves good performance to approximate TrustSet, where only small gap exists when labeled dataset becomes larger. In Cifar10-imb dataset, similarly, when labeled dataset is limited, BRAL-T achieves similar accuracy compared with GradNd. When the size of labeled dataset becomes larger, the accuracy difference performs to be acceptable larger. As a conclusion, the RL policy in BRAL-T achieves good performance to approximate ground truth TrustSet selection.

\end{document}

%% file: math_commands.tex
\usepackage{amsmath,amsfonts,bm}

\def\eqref#1{equation~\ref{#1}}

\def\1{\bm{1}}

\DeclareMathAlphabet{\mathsfit}{\encodingdefault}{\sfdefault}{m}{sl}
\SetMathAlphabet{\mathsfit}{bold}{\encodingdefault}{\sfdefault}{bx}{n}

\DeclareMathOperator*{\argmin}{arg\,min}